\documentclass[11pt,a4paper]{article}
\usepackage[hyperref]{emnlp2018}
\usepackage{times}
\usepackage{latexsym}
\usepackage{amsmath}
\usepackage{graphicx}
\usepackage{url}

\aclfinalcopy 

\title{Parameterized Convolutional Neural Networks for Aspect Level Sentiment Classification}

\author{Binxuan Huang\\
School of Computer Science\\
Carnegie Mellon University \\
  {\tt binxuanh@cs.cmu.edu} \\\And
  Kathleen M. Carley \\\
School of Computer Science\\
Carnegie Mellon University \\
  {\tt kathleen.carley@cs.cmu.edu} \\}

\date{}
\begin{document}
\maketitle
\begin{abstract}
We introduce a novel parameterized convolutional neural network for aspect level sentiment classification. Using parameterized filters and parameterized gates, we incorporate aspect information into convolutional neural networks (CNN). Experiments demonstrate that our parameterized filters and parameterized gates effectively capture the aspect-specific features, and our CNN-based models achieve excellent results on SemEval 2014 datasets. 
\end{abstract}

\section{Introduction}

Continuous growing of user generated text in social media platforms such as Twitter drives sentiment classification increasingly popular.
The goal of sentiment classification is to detect whether a piece of text expresses a positive, a negative, or a neutral sentiment \cite{rosenthal2017semeval}. The majority of the literature focuses on general sentiment analysis (document level), not involving a specific topic or entity. When there are multiple aspects about an entity in a sentence, it is hard to determine the sentiment as a whole.

Differing from general sentiment classification, aspect level sentiment classification identifies opinions from text about specific entities and their aspects \cite{pontiki2015semeval}. For example, given a sentence ``great food but the service was dreadful", the sentiment polarity about aspect ``food" is positive while the sentiment polarity about ``service" is negative. If we ignore the aspect information, it is hard to determine the sentiment for a target aspect, which accounts for a large portion of sentiment classification error \cite{jiang2011target}.

Recently, machine learning based approaches are becoming popular for this task. Representative approaches in literature include Support Vector Machine (SVM) with manually created features \cite{jiang2011target,wagner2014dcu} and neural network based models \cite{tang2015effective,wang2016attention,huang2018aspect}. Because of neural networks' capacity of learning representations from data without feature engineering, they are of growing interest for this natural language processing task. The mainstream neural methods are either based on long short-term memory \cite{hochreiter1997long} or memory networks \cite{sukhbaatar2015end}. None of them are using convolutional neural networks (CNN), which are good at capturing local patterns.

In the present work, we propose two simple yet effective convolutional neural networks with aspect information incorporated. The overall architecture differs significantly from previous work. Specifically, we design two novel neural units that take target aspects into account. One is parameterized filter, the other is parameterized gate. These units both are generated from aspect-specific features and are further applied on the sentence. Our experiments show that our two model variants work surprisingly well on this type of task. 
\section{Related Work}
Aspect level sentiment classification is a branch of sentiment classification \cite{pang2002thumbs,wang2012baselines}. It aims at identifying the sentiment polarity of one aspect target in a context sentence. 

One typical early work tries to identify the aspect level sentiment polarity based on predefined language rules \cite{nasukawa2003sentiment}. Nasukawa and Yi first perform dependency parsing on sentences. Then rules are applied to determine the sentiment about aspects. Standard machine learning algorithms like SVM are also widely used on this task. Jiang et al. create several target-dependent features, then they feed these target-dependent features with content features into an SVM classifier.

In recent years, aspect level sentiment classification is dominated by neural network based approaches. The majority of published works rely on the architecture of long short-term memory (LSTM) \cite{tang2015effective}. \newcite{wang2016attention} use an attention vector generated from aspect embedding to better capture the important parts in sentences. \newcite{tay2017learning} introduce a word-aspect fusion operation to learn associative relationships between aspects and sentences. \newcite{huang2018aspect} use an attention-over-attention layer to further improve the performance.

Another type of neural architectures known as memory network \cite{sukhbaatar2015end} has also been used in this task. \newcite{tang2016aspect} takes an aspect term as a query sent to external memory. Their model consists of multiple computational layers. Each layer is an attention model. One recent work Dyadic MemNN \cite{tay2017dyadic} places associative layers on top of memory networks to improve the performance.

The overall architecture in this paper differs significantly with all these previous works. To the best of our knowledge, this paper is the first attempt using convolutional neural networks \cite{kim2014convolutional,huang2017predicting} for aspect level sentiment classification.

\section{Parameterized Convolutional Neural Networks}

In this section, we introduce our method for aspect level sentiment classification, which is based on convolutional neural networks. We first describe CNN for general sentiment classification, then we introduce our two model variants Parameterized Filters for Convolutional Neural Networks (PF-CNN) and Parameterized Gated Convolutional Neural Networks (PG-CNN).

\subsection{Problem Definition}
In aspect level sentiment classification, we are given a sentence $s=[w_1,w_2,$ $...,w_i,...,w_n]$ and an aspect target $t=[w_i,w_{i+1},...,w_{i+m-1}]$. The goal is to classify whether the sentiment towards the aspect in the sentence is positive, negative, or neutral.

\begin{figure*}[!h]
    \centering
    \includegraphics[width=1.0\textwidth]{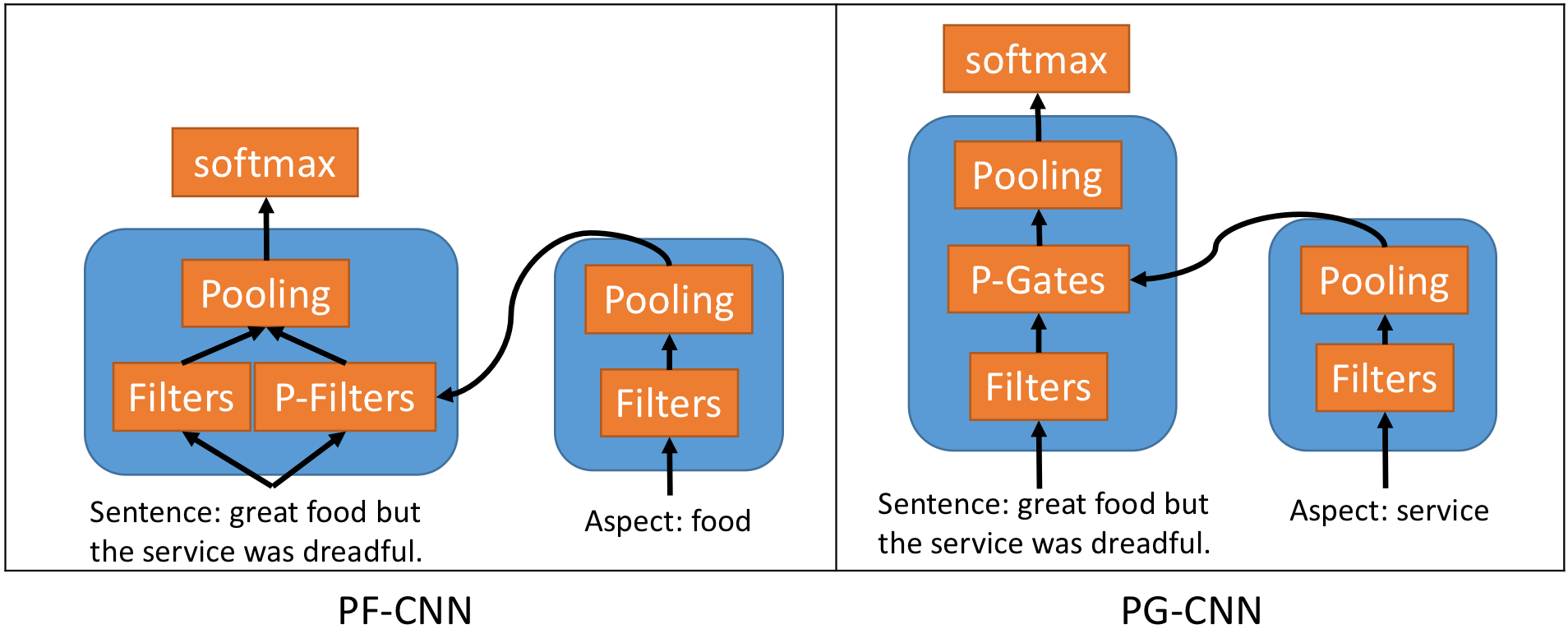}
      \vspace{-1cm}
    \caption{The overall architectures of PF-CNN and PG-CNN.}
    \vspace{-0.5cm}
    \label{arch}
\end{figure*}
\subsection{Convolutional Neural Networks}
We first briefly describe convolutional neural networks (CNN) for general sentiment classification \cite{kim2014convolutional}. Given a sentence $s=[w_1,w_2,$ $...,w_i,...,w_n]$, let $v_i\in R^k$ be the word vector for word $w_i$. A sentence of length $n$ can be represented as a matrix $\mathbf s =[v_1,v_2,...,v_n]\in R^{n\times k}$. A convolution filter $w\in R^{h\times k}$ with width $h$ is applied to the word matrix to get high-level representative features. Specifically, for a word window $v_{i:{i+h-1}}\in R^{h\times k}$, a feature $c_i$ is generated by

\vspace{-0.3cm}
\begin{equation}
\label{1}
c_i=f(w\odot v_{i:{i+h-1}}+b)
\end{equation}
where $\odot$ represents element-wise product, $b\in R$ is a bias term and f is a non-linear function. Sliding the filter window from the beginning of the word matrix till the end, we get a feature map $\mathbf c\in R^{n-h+1}$.
\vspace{-0.3cm}
\begin{equation}
\mathbf c=[c_1,c_2,...,c_{n-h+1}]
\end{equation}
After that, a pooling operation is applied over the feature map to get one single general sentiment feature $\theta_g$ in each map. We use max pooling in the CNN for sentences.

\vspace{-0.3cm}
\begin{equation}\theta_g=pooling(\mathbf c) \end{equation}
We denote this process as $\theta_g = CNN_g(\mathbf s;w,b)$.
Using $d$ such convolutional filters, we can get a general sentiment feature vector $ \Theta_g \in R^{d}$ without information from aspect terms.

\subsection{Parameterized Filters}
Standard convolutional neural networks do not consider information from aspect terms. Herein, our first model variant overcomes this issue by parameterizing filters using aspect terms. We call it Parameterized Filters for Convolutional Neural Networks (PF-CNN). The overall architecture is shown in the left of Figure \ref{arch}.

Formally, given the aspect term with length $m$, $t=[w_i,w_{i+1},...,w_{i+m-1}]$ and the corresponding embedding matrix $\mathbf t=[v_i,v_{i+1},...,v_{i+m-1}]\in R^{m\times k}$, we first use another $CNN_t$ to extract one single feature $\theta_t$ from $\mathbf t$.
\begin{equation}
\theta_t = CNN_t(\mathbf t; w_t,b_t)
\end{equation}
where $w_t\in R^{h_t\times k}$, $b_t$ are the convolutional filter, bias term for $CNN_t$. $h_t$ is the width of filters applied on aspect targets. With $h_s\times k$ such filters and bias terms, we can get a feature matrix $\Theta_t\in R^{h_s\times k}$, where $h_s$ is the filter width applied on sentences. We use average pooling in the $CNN_t$ for aspects.

In the next step, $ \Theta_t$ is further used as a convolutional filter applied on the sentence.
\begin{equation}
\label{tcnn}
\theta_s = CNN_s(\mathbf s; \mathbf \Theta_t,b_s)
\end{equation}
Using such $d$ parameterized filters, we get the aspect-specific features $\Theta_s\in R^{d}$ with target term information. We further concatenate the targeted feature vector with general sentiment features as the final classification features $\Theta = [\Theta_g,\Theta_s]$.

\subsection{Parameterized Gates}
The second model variant we designed is called Parameterized Gated Convolutional Neural Networks (PG-CNN). The overall architecture is shown in the right of Figure \ref{arch}.

Similar with PF-CNN, PG-CNN also utilizes a convolutional neural network to extract feature $\Theta_t$ from aspect terms, which instead is used as a gate \cite{dauphin2016language} in the CNN applied on the sentence. The key step of PG-CNN is described in equation (\ref{pgcnn}). 
\begin{equation}
\label{pgcnn}
c_i = (w\odot v_{i:i+h-1}+b)\cdot \sigma (\Theta_t \odot v_{i:i+h-1} +b) 
\end{equation}
Instead of using a non-linear function $f$ in equation (\ref{1}), we use a gate $\sigma (\Theta_t \odot v_{i:i+h-1} +b)$ to control how much information passing to the next layer, where $\sigma(\cdot)$ is sigmoid function. For each general filter applied on the sentence, one parameterized gate is generated from the aspect.

After that, we generate the final classification feature $\Theta$ in the same way as standard CNN.

\subsection{Final Classification}
We feed the final classification feature into a linear layer to project $\Theta$ into the space of targeted classes:
\begin{equation}
x = W_l\cdot \Theta+b_l
\end{equation}
where $W_l$ and $b_l$ are the weight matrix and bias. Following the linear layer, we use a softmax layer to compute the probability of class $c$.
\begin{equation}
P(y=c|x)=\frac{exp(x_c)}{\sum_{i\in C}exp(x_i)}
\end{equation}

\subsection{Model training}
We train our model to minimize the cross-entropy loss function with $L_2$ regularization:
\begin{align*}
loss = -\sum_{(s,t)}\sum_{c\in C} I(y=c)logP(y=c|s,a)+\lambda||p||^2
\end{align*}
where $I(\cdot)$ is the indicator function and $p$ is the set of all parameters in the convolutional layers and linear layer.

\section{Experiments}
\subsection{Experiments Setting}
\textbf{Dataset}\\
We adopt one widely used dataset from SemEval 2014 Task 4 \cite{pontiki2014semeval}. It contains two domain-specific datasets for laptops and restaurants. Each data point is a pair of a sentence and an aspect term. Experienced annotators tagged each pair with sentiment polarity. Following recent work \cite{tay2017learning}, we take 500 training instances as development set\footnote{The splits can be found at \href{https://github.com/vanzytay/ABSA_DevSplits}{https://github.com/vanzytay/ABSA\_DevSplits}.}. Unfortunately, many works have not mentioned the usage of development set \cite{wang2016attention,ma2017interactive}.

\begin{table}[]
\centering
\label{data}
\begin{tabular}{|l|l|l|l|}
\hline
Dataset          & Positive & Neutral & Negative \\ \hline
Laptop-Train     & 767      &   373   & 673      \\ \hline
Laptop-Dev     & 220      & 87     & 193      \\ \hline
Laptop-Test      & 341      & 169     & 128      \\ \hline
Restaurant-Train & 1886     & 531     & 685      \\ \hline
Restaurant-Dev & 278     & 102     & 120      \\ \hline
Restaurant-Test  & 728      & 196     & 196      \\ \hline
\end{tabular}
\caption{Statistics of the datasets. }
\end{table}

\noindent \textbf{Hyperparameters and Training}\\
We use rectifier as non-linear function $f$ in the $CNN_g$, $CNN_t$ and sigmoid in the $CNN_s$, filter window sizes of $1,2,3,4$ with 100 feature maps each, $l_2$ regularization term of $0.001$ and mini-batch size of 25. Parameterized filters and  gates have the same size and number as normal filters. They are generated uniformly by CNN with window sizes of $1,2,3,4$, eg. among 100 parameterized filters with size 3, 25 of them are generated by aspect CNN with filter size 1, 2, 3, 4 respectively. The word embeddings are initialized with 300-dimensional Glove vectors \cite{pennington2014glove} and are fixed during training. For the out of vocabulary words we initialize them randomly from uniform distribution $U(-0.01,0.01)$. We apply dropout on the final classification features of PG-CNN. The dropout rate is chosen as 0.3.

Training is done through mini-batch stochastic gradient descent with Adam update rule \cite{kingma2014adam}. The initial learning rate is 0.001. If the training loss does not drop after every three epochs, we decrease the learning rate by half. We adopt early stopping based on the validation loss on development sets.

\subsection{Results}

\begin{table}[]
\centering
\begin{tabular}{|l|l|l|l|l|}
\hline
          & \multicolumn{2}{l|}{Laptops} & \multicolumn{2}{l|}{Restaurants} \\ \hline
Model     & 3-way        & Binary        & 3-way          & Binary          \\ \hline
TD-LSTM   & 62.38        & 79.31         & 69.73          & 84.41           \\ \hline
AT-LSTM   & 65.83        & 78.25         & 74.37          & 84.74           \\ \hline
ATAE-LSTM & 60.34        & 74.20         & 70.71          & 84.52           \\ \hline
AF-LSTM   & 68.81        & 83.58         & 75.44          & 87.78           \\ \hline 
CNN & 68.65        & 85.50         & 77.95          & 89.50          \\ \hline 
PF-CNN     & \textbf{70.06}            &    \textbf{86.35}           &               \textbf{79.20} &   {90.15}              \\ \hline
PG-CNN     & {69.12}            &    {86.14}           &               {78.93} &   \textbf{90.58}              \\ \hline
\end{tabular}
\caption{Comparisons results with baselines. We use accuracy to measure the performance. Performances of baselines are cited from \cite{tay2017learning}.}
\vspace{-0.4cm}
\label{compare}
\end{table}
We use accuracy metric to measure the performance. To show the effectiveness of our model, we compare it with several baseline methods. We list them as follows:


\textbf{TD-LSTM} uses two LSTM networks to model the preceding and following contexts surrounding the aspect term. The last hidden states of these two LSTM networks are concatenated for predicting the sentiment polarity \cite{tang2015effective}.

\textbf{AT-LSTM} combines the sentence hidden states from a LSTM with the aspect term embedding to generate the attention vector. The final sentence representation is the weighted sum of the hidden states \cite{wang2016attention}.

\textbf{ATAE-LSTM} further extends AT-LSTM by appending the aspect embedding into each word vector \cite{wang2016attention}.

\textbf{AF-LSTM} introduces a word-aspect fusion attention to learn associative relationships between aspect and context words \cite{tay2017learning}.

\textbf{CNN} uses the architecture proposed in  \cite{kim2014convolutional} without explicitly considering aspect. We use filter window sizes of 1,2,3,4 with 100 maps each. Dropout rate is chosen as 0.5. Early stopping based on validation accuracy is applied.

Our two models achieve the best performance when compared to these baselines as shown in Table \ref{compare}, which shows that our proposed neural units effectively captures the aspect-specific features. Compared to one recently proposed model AF-LSTM, our method achieve 2\%-5\% improvements. 
 Surprisingly, a vanilla CNN works quite well on this problem. It even beats these well-designed LSTM models, which further proves that using CNN-based methods is a direction worth exploring.
 
 \subsection{Case Study \& Discussion}
 Compared to a vanilla CNN, our two model variants could successfully distinguish the describing words for corresponding aspect targets. In the sentence ``the appetizers are ok, but the service is slow'', a vanilla CNN outputs the same negative sentiment label for both aspect terms ``appetizers'' and ``service'', while PF-CNN and PG-CNN successfully recognize that ``slow'' is only used for describing ``service'' and output neutral and negative labels for aspects ``appetizers'' and ``service'' respectively.
 In another example ``the staff members are extremely friendly and even replaced my drink once when i dropped it outside'', our models also find out that positive and neutral sentiment for ``staff'' and ``drink'' respectively.
 
One thing we notice in our experiment is that a vanilla CNN ignoring aspects has comparable performance with some well-designed LSTM models in this aspect-level sentiment classification task.
For a sentence containing multiple aspects, we assume the majority of the aspect-level sentiment label is the sentence-level sentiment label. Using this labeling scheme, in the restaurant data, 1034 out of 1117 test points have the same sentence-level and aspect-level labels. Thus, a sentence-level classifier with accuracy 75\% also classifies 70\% aspect-labels correctly. A similar observation was made for the laptop dataset as well.
Probably this is the reason why a vanilla CNN has comparable performance on these two datasets. For future research, a more balanced dataset would be helpful to overcome this issue.

\section{Conclusion}
We propose a novel method for aspect level sentiment classification. We introduce two novel neural units called parameterized filter and parameterized gate to incorporate aspect information into the convolutional neural network architecture. Comparisons with baseline methods show our model effectively learns the aspect-specific sentiment expressions. Our experiments demonstrate a significant improvement compared to multiple strong neural baselines.

To the best of our knowledge, our model is the first attempt using convolutional neural networks solving this problem. We hope this work could inspire future research exploring in this direction. It is also interesting to see whether such parameterized CNN architecture could benefit other natural language processing tasks involving text pairs like question answering task.

\section*{Acknowledgments}
We would like to thank the reviewers for their helpful comments that greatly improved the
article. We would also like to thank Sumeet Kumar for his valuable suggestions. 
\bibliographystyle{acl_natbib}
\bibliography{./reference}
\end{document}